\def\year{2020}\relax
\documentclass[letterpaper]{article} % DO NOT CHANGE THIS
\usepackage{aaai20}  % DO NOT CHANGE THIS
\usepackage{times}  % DO NOT CHANGE THIS
\usepackage{helvet} % DO NOT CHANGE THIS
\usepackage{courier}  % DO NOT CHANGE THIS
\usepackage[hyphens]{url}  % DO NOT CHANGE THIS
\usepackage{graphicx} % DO NOT CHANGE THIS
\usepackage{algorithm}
\usepackage{algorithmic}
\usepackage{amsmath}
\usepackage{booktabs}
\usepackage{diagbox}
\usepackage[T1]{fontenc}

\usepackage{graphicx}  % DO NOT CHANGE THIS
\title{Learning to Reason in Round-based Games: Multi-task Sequence Generation for Purchasing Decision Making in First-person Shooters}

\author{Yilei Zeng\thanks{\enspace Equal contributions.}, Deren Lei\footnotemark[1], Beichen Li\footnotemark[1], Gangrong Jiang\footnotemark[1], Emilio Ferrara, Michael Zyda\\
University of Southern California\\
\{yilei.zeng, derenlei, beichen, gjiang, emiliofe, zyda\}@usc.edu\\
}

\begin{document}

\maketitle

\begin{abstract}

Sequential reasoning is a complex human ability, with extensive previous research focusing on gaming AI in a single continuous game, round-based decision makings extending to a sequence of games remain less explored. Counter-Strike: Global Offensive (CS:GO), as a round-based game with abundant expert demonstrations, provides an excellent environment for multi-player round-based sequential reasoning. In this work, we propose a Sequence Reasoner with Round Attribute Encoder and Multi-Task Decoder to interpret the strategies behind the round-based purchasing decisions. We adopt few-shot learning to sample multiple rounds in a match, and modified model agnostic meta-learning algorithm Reptile for the meta-learning loop. We formulate each round as a multi-task sequence generation problem. Our state representations combine action encoder, team encoder, player features, round attribute encoder, and economy encoders to help our agent learn to reason under this specific multi-player round-based scenario. A complete ablation study and comparison with the greedy approach certify the effectiveness of our model. Our research will open doors for interpretable AI for understanding episodic and long-term purchasing strategies beyond the gaming community.
%expert demonstrations has been proven essential for training gaming agents
%What are the strategies of professional players when making decisions on purchasing in round-based game. 
%We study the problem of developing artificial intelligent agents for round-based games, which is a crucial part of modern games. More specifically, we introduce a new round-based game dataset based on Counter-Strike: Global Offensive (CS:GO). To the best of our knowledge, this is the first dataset that targets learning to reason in round-based games. Compared to other widely used game reasoning datasets, its dynamically changing scenario is more complex and each scenario contains a long consecutive of episodes. We cast this task into the learning-to-learn framework. Our proposed method formulates each episode/round as a multi-task sequence generation problem that leverages Deep Reinforcement Learning (DRL). A complete ablation study and comparison with the greedy approach certify the effectiveness of our model. Furthermore, we introduce a round attribute encoder and demonstrate that it can help our agent learn to reason under this specific round-based scenario. \footnote{\small Code and data will be released.}
\footnote{\small Code and data is available at \url{https://github.com/derenlei/CS_Net}.}

\end{abstract}

\section{Introduction}
%Games as a research framework with recent AI milestones achieved in Atari games, Alpha Go, Alpha zero, and Poker have attracted a vast amount of attention. 

Over the last couple of years, we have seen the gaming AI community moving towards training agents in more sophisticated games, like Doom \cite{kempka2016vizdoom}, StarCraft \cite{lin2017stardata}, Minecraft \cite{guss2019minerl}, and Dota2 \cite{openai2019dota}. These online games are match-based, fast-paced, highly strategic, and involve adversarial real-time battles. 

%Artificial Intelligence (AI) in video games is a long-standing research area. Deep learning models have tackled tasks such as StarCraft \cite{lin2017stardata}, , Dota2 \cite{openai2019dota} and Doom \cite{kempka2016vizdoom}. They start to reason in complex and continuous environments that capture the nature of the real world.
%However, the reasoning of gaming AI agents are contained within a single continuous match as one complete game. The sequential reasoning in round-based games are ignored.

However, existing studies are restricted to treating the complete continuous game as a single task which would not generalize to the cases of round-based games. We define round-based game as a meta game that can be decomposed into multiple games which can be independent and with different rules. An round-based game reasoning example could be: A professional Dota2 player plays five games in total, but already lose two games in a row. How would he play the third one? Will he choose an aggressive strategy? Such round-based game reasoning is a fundamental problem of building video game AI, as its strategy can also be applied to continuous game cases. For example, the death of a player's character in DOTA2 can be considered as a round or a micro-game which contributes non-equally to the entire meta-game. Utilizing information from a round-based level could be crucial to the complete game reasoning process. 

In this work, we are particularly interested in tackling the new challenges of round-based games by proposing a simple scenario: each round only requires one action to receive a result, and each round contributes equally to the entire meta-game. To this end, we introduce a round-based dataset collected from CS:GO professional match replays, which consists of 5167 matches, and each match contains a maximum of 30 rounds and exactly 10 players for two sides. Winning the entire game requires to win at least 16 rounds.

The agent learns the reasoning behind the player's weapon purchasing decisions at the beginning of each round, given the observation of previous rounds' match statistics together with the current round weapon information described in Section~\ref{Sec:Dataset}. We aim to build a human centered AI that learn to reason from human decision makings and in return help interpret the process rather than achieving the best in game performance.

%We use $F_1$ score not only as the evaluation metrics to measure whether our model can predict the purchasing decision correctly but also as the reward provided to the agent from the environment. Each team could develop several multi-round economy strategies that deliberately let go some disadvantaged rounds temporarily to save money and build up comparative advantages for future rounds. As for each game, each player has its preference of weapon purchasing. Also, its financial status profoundly impacts the player's purchasing policy in each round. Due to the complexity and diversity of each player's attributes (policy), we cast it into the learning-to-learn framework. For each game, the first few rounds are observed, and the model is required to predict well in later rounds. This few-shot learning setting relieves us from the difficulties of tedious feature engineering and training attribute representations from insufficient data. Such a framework is capable of dealing with circumstances where a team adjusts its strategy in different matches, or a player changes its style to fit in the team. It addresses these issues by requiring the model to adapt to an initialization that can quickly approach the optimum after a few shots of updates.

%\subsection*{Contributions}

We propose three approaches to deal with such reasoning challenge: 

\begin{itemize}
    \item \textit{Greedy Algorithm}, this model buys the utmost affordable weapon sequence during each round.

    \item \textit{Sequence Reasoner}, since a player buys a sequence of weapons for each round and each weapon belongs to one of three types, gun, grenade, or equipment, we consider this task as a multi-task sequence generation problem with pre-trained the weapon embedding from the context of weapon sequence.
    
    \item \textit{Sequence Reasoner with Round Attribute Encoder}, the model encodes the player's previous round history through Round Attribute Encoder (RAE) into an auxiliary round attribute for the Sequence Reasoner.  
    % \item \textit{Sequence Reasoner with Round Attribute Encoder}, the model encodes player's previous round history final weapons through Round Attribute Encoder (RAE) as extra round attribute input to the Sequence Reasoner. We perform extensive experiments to investigate their performances: our Deep Learning framework performances are reasonable and higher than \textit{Greedy Algorithm}, but still far below original professional players' performances. Thus, we believe that the proposed CS:GO weapon purchasing dataset can be an important new benchmark for round-based AI reasoning. Besides, our ablation studies demonstrate the effectiveness of RAE.
\end{itemize}
Extensive experiments demonstrate the effectiveness of our proposed third method. However, the result is still not close to the original professional player's level. Thus, we believe that the proposed CS:GO weapon purchasing dataset can be an important new benchmark and our model shades light for future works on round-based AI reasoning.
%We perform extensive experiments to investigate model performances: our ablation studies demonstrate the effectiveness of the proposed RAE. Our Deep Learning framework shows more interpretive and better performances than \textit{Greedy Algorithm}, but not yet reached original professional players' level.  Thus, we believe that the proposed CS:GO weapon purchasing dataset can be an important new benchmark and shades light for future works on round-based AI reasoning.

\section{Related Work}

\subsubsection{Learning to Learn \& Few-Shot Learning}
Existing learning to learn or meta-learning studies \cite{hochreiter2001learning,thrun2012learning} mainly focus on supervised learning problems. A particularly challenging problem is learning with few training examples, i.e. few-shot learning. Generally, few-shot learning datasets contain several tasks, and for each task, there are a limited number of examples with supervised information. Few-shot learning algorithms can improve on new tasks using provided supervision. Within the learning process, the meta knowledge is extracted by a meta-learner, which learns to generalize the meta knowledge on each specific task. \citeauthor{vinyals2016matching} \shortcite{vinyals2016matching} uses Matching Networks with attention and memory to enable rapid learning. \citeauthor{snell2017prototypical} \shortcite{snell2017prototypical} propose Prototypical Networks. While \citeauthor{ravi2016optimization} \shortcite{ravi2016optimization} use an LSTM-based meta-learning to learn an update rule for training a neural network learner, Model-Agnostic Meta-Learning (MAML) \cite{finn2017model} learns a good model parameter initialization that can quickly adapt to similar tasks. Reptile \cite{nichol2018reptile} is a first-order approximation of MAML, which is remarkably simple and performs similarly well. In this study, we adapt Reptile to our framework.

%\subsubsection{Dynamically Changing Environment}
%The ability to quickly adapt from limited experience in dynamically changing environments is an important topic for Artificial Intelligence. Early works approach this problem by context detection \cite{da2006dealing} and tracking \cite{sutton2007role}. These methods that fine-tune the policy in changing environments suffer from inefficient sample problems. \citeauthor{al2017continuous} \shortcite{al2017continuous} tackles this problem as a multi-task problem \cite{caruana1997multitask} since a non-stationary environment can be seen as a sequence of stationary ones. This approach is widely used for online learning problems \cite{nagabandi2018learning,park2018meta,nagabandi2018deep}. In our introduced round-based task, each round can be considered as an changed environment. %Thus, the dynamically changing environment is more complex in our scenario.

\subsubsection{Gaming Machine Learning Datasets}
%Applications of Machine Learning in games have been widely researched. 
Datasets and environments are crucial for facilitating gaming machine learning research by serving as benchmarking platforms for new methods. STARDATA \cite{lin2017stardata} is currently the largest StarCraft AI Research Dataset, dedicated to real-time strategy (RTS) games research. \citeauthor{hu2019hierarchical} \shortcite{hu2019hierarchical} provides a simpler RTS environment that focus more on language instruction as macro-actions. For first-person shooter (FPS) games, VizDoom \cite{kempka2016vizdoom} provides an environment for 3D visual Reinforcement Learning. MineRL \cite{guss2019minerl} introduces a large-scale, simulator-paired dataset of human demonstrations of sandbox game MineCraft. The Atari games are popular for RL methods evaluation, and The Atari Grand Challenge Dataset \cite{kurin2017atari} has catalyzed the research. These studies focus on continuous environments without interruptions. In our work, we introduce a novel round-based gaming dataset based on CS:GO. Each round can be considered as an independent gaming episode that equally contributes to the match where a multi-round meta-strategies exist. Players can strategically lose some gaming episodes in exchange for winning a long-term goal.

\section{Task}

\subsection{Few-shot Learning}
%We use $F_1$ score not only as the evaluation metrics to measure whether our model can predict the purchasing decision correctly but also as the reward provided to the agent from the environment.
Each team could develop several multi-round economy strategies that deliberately let go of some disadvantaged rounds temporarily to save money and build up comparative advantages for future rounds. As for each game, each player has its own preference of weapon purchasing. The financial status profoundly impacts the player's purchasing policy in each round. Due to the complexity and diversity of each player's attributes (policies), we cast the task into the learning-to-learn framework. For each game, the first few rounds are observed, and the model learns to predict later rounds. This few-shot task setting brings more opportunities for agents to learn players' dynamic attributes during inference and challenge the agents to learn more generalized policies that can quickly be adapted to current players after some observations.

\subsection{Problem Formulation}
We treat each match as a separate data point. Each match consists 10 different tasks from 10 players' perspectives. Since each player has its own preference for weapons, we formulate the problem into a few-shot learning task to foster the agent to capture the preference from the few support shots. 
%Few-shot learning examines an agent's ability to quickly adapt to a new task given only a few training examples. 
For each match $M_i$, each player $i$ go through $j$ rounds, rounds are noted as $R_{i,j},j\in[ \,1,n_{i}] \,$, $M_i = \{R_{i,1}, R_{i,2}, ..., R_{i,n_{i}}\}$. We use the first $K$ rounds as K-shot training examples. The model adjusts on the $K$ shots (support set) and is asked to behave well on other $n_{i}-K$ rounds (target set).

We formulate the problem in a reinforcement learning setup. At the beginning of each round, an agent (a player) estimates the states and takes a single action that stands for a weapon purchasing set. The state of the agent includes weapons and money of all players.
%The environment is partially observable since the real-time attributes of teammates and enemies are not known.

We introduce the formulation for a match $M_i$ and for simplicity, we omit the subscript $i$. For the match $M$, the agent's possessions at the $j$-th round is composed of the weapons it owns $X_j = \{x_{j,1}, x_{j,2}, \dots, x_{j,m_j}\}$ and money $c_j$. At round $j$, the history information of this round $H_j=\{E_1, E_2, \dots, E_{j-1}\}$ contains empirical information of past rounds. The empirical information of $j$-th round $E_j$ consists of final weapons after purchasing $X'_j=\{x'_{j,1}, x'_{j,2}, \dots\}$ and performance score $s_j$ . For the $j$-th round, given an agent's own weapons $X^s_j$, team's weapons $X^t_j=\{X^t_{j,1}, X^t_{j,2}, X^t_{j,3}, X^t_{j,4}, X^t_{j,5}\}$ and opponent's weapons $X^o_j=\{X^o_{j,1}, X^o_{j,2}, X^o_{j,3}, X^o_{j,4}, X^o_{j,5}\}$, along with history information $H_j=\{E_1, E_2, \dots, E_{j-1}\}$ from past rounds and all players' money, the agent needs to properly generate the action $A_j$ to approach the label $\hat{A_j}$.

\section{Dataset}
\label{Sec:Dataset}

\begin{table}[t!]
\centering
\scalebox{0.95}{
  \begin{tabular}{l | c}
    \toprule
    \textbf{Attribute}     & \textbf{Description}     \\
    \midrule
    account & current cash\\
    cash spent & cash the player has spent this round\\
    weapons & all weapons held by this player\\
    items value & sum of current items' prices\\
    performance score & player's score in the scoreboard\\
    \bottomrule
  \end{tabular}
  }
  \caption{\textbf{Description of the extracted information of a player for each round.} Performance score is described in Section~\ref{subsec:replay}.}
  \label{tab:data}
\end{table}

We aim to build a dataset dedicated for round-based game reasoning. To build our dataset, we collect professional CS:GO players' match-replays during 2019.

%This few-shot learning setting relieves us from the difficulties of tedious feature engineering and training attribute representations from insufficient data. Such a framework is capable of dealing with circumstances where a team adjusts its strategy in different matches, or a player changes its style to fit in the team. It addresses these issues by requiring the model to adapt to an initialization that can quickly approach the optimum after a few shots of updates.

\subsection{Parsing Replays}
\label{subsec:replay}
We design a systematic procedure to process the replays. First, we parse the replays using the demofile parser\footnote{ \url{https://saul.github.io/demofile/index.html}}. We then filter out anomalous data to ensure quality. For all the replays, we extract all information related to weapon purchasing. Specifically, we capture players' weapon pickup and weapon removal actions. We also extract each player's state 3 times each round, including round start, weapon purchasing period end, and round end. Table~\ref{tab:data} shows detailed description for the extracted states. The performance score is provided by the CS:GO in-game scoreboard, which is based on the player's kill and bomb planted/defused. We use the normalized score in our Round Attribute Encoder for past rounds' encoding.

After parsing, we convert them into structured JSON format. Data cleaning is subsequently performed to ensure data quality. We drop out the data with inconsistencies on weapon purchased and money spent. We then obtain consistent data of 5167 matches. We randomly shuffle the matches and split the data into training, development, and test set in the ratio 8:1:1. The training, development, and test set consist of 4133, 517, and 517 matches, respectively.

\subsection{Statistics}
\begin{table}[t!]
\centering
\scalebox{0.85}{
  \begin{tabular}{l | ccccc}
    \toprule
    \backslashbox{\textbf{Type}}{\textbf{Count}}  & \textbf{0} & \textbf{1} & \textbf{2} & \textbf{3} & \textbf{4}   \\
    \midrule
    Gun & 35.9\% & 61.6\% & 2.4\% & 0.1\% & 0\%   \\
    Grenade & 19.4\% & 12.6\% & 14.6\% & 16.4\% & 37.0\%   \\
    Equipment & 38.3\% & 50.3\% & 10.7\% & 0.7\% & 0\%   \\
    \bottomrule
  \end{tabular}
  }
  \caption{\textbf{The distribution of purchasing action count for each type of weapon.} Note that gun and equipment purchased 4 times in one purchasing sequence are very rare so they are rounded down and visualized as 0\% in this table.}
  \label{tab:action-count}
\end{table}

In CS:GO, there are 44 different weapons in total, including 34 guns, 6 grenades and 4 equipments. For guns, there are 6 different types: pistols, shotguns, SMGs, Automatic Rifles, LMGs and Sniper Rifles. Equipment includes helmet, vest, defuse kit and Zeus x27. In order to get high-level representations for their intrinsically diverse attributes, we use a self-supervised learning method to train the embedding of weapons. We treat guns, grenades and equipment as three types and perform generation separately in our model.

Table~\ref{tab:action-count} shows the frequency distribution of three categories of purchasable items, i.e., gun, grenade, and equipment, in the purchasing sequences per round. Table ~\ref{tab:action-count} does not contain information about the the exact position within a sequence. We sort  all the weapon sequences in the order of gun, grenade, and equipment based on our human prior knowledge.

Although a player can only carry 1 primary gun and 1 pistol, there are some cases that players buy more than 2 guns, probably for their teammates. As in this paper, we only consider studying the task as a single agent round-based problem, we do not remove these cases and leave learning collaborative purchasing to future work. The player can carry 4 grenades and 1 of each type of equipment at maximum so the maximum action length for purchasing is 4. Each row shows for each type of weapon when in the purchasing sequence it is likely to be purchased.

\section{Methodology}\label{method}
\begin{figure*}
    \centering
    \includegraphics[width=15cm]{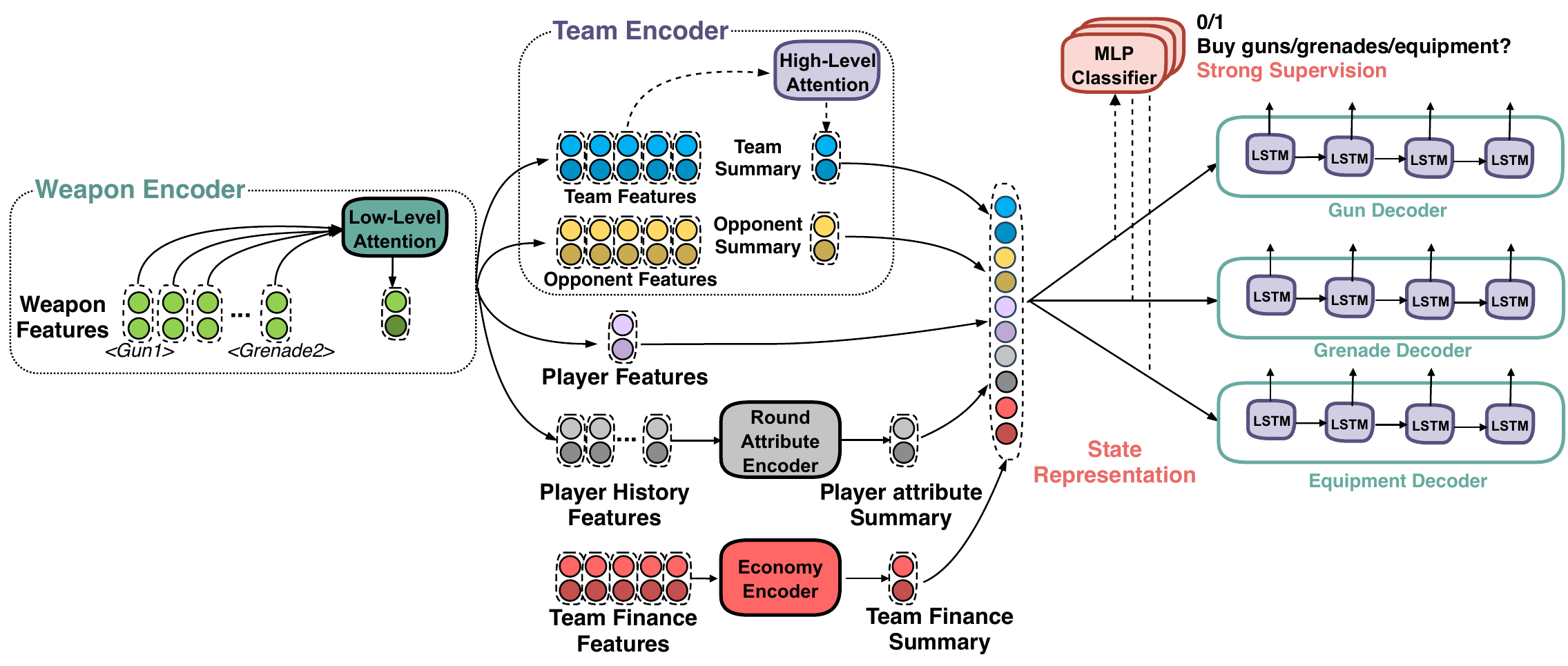}
    \caption{\textbf{Model Architecture}. Weapon encoder generates a weapon summary based on a set of weapon features of a player. Team encoder encodes a set of weapon summaries into a single team summary. Round attribute encoder encodes the player's information of previous rounds. The state representation is then fed into three LSTMs separately. Three gate classifiers are trained with strong supervision to determine if the generation of a certain type of action is suitable.}
    \label{fig:model}
\end{figure*}{}

\begin{figure}[h!]
    \centering
    \includegraphics[width=8cm]{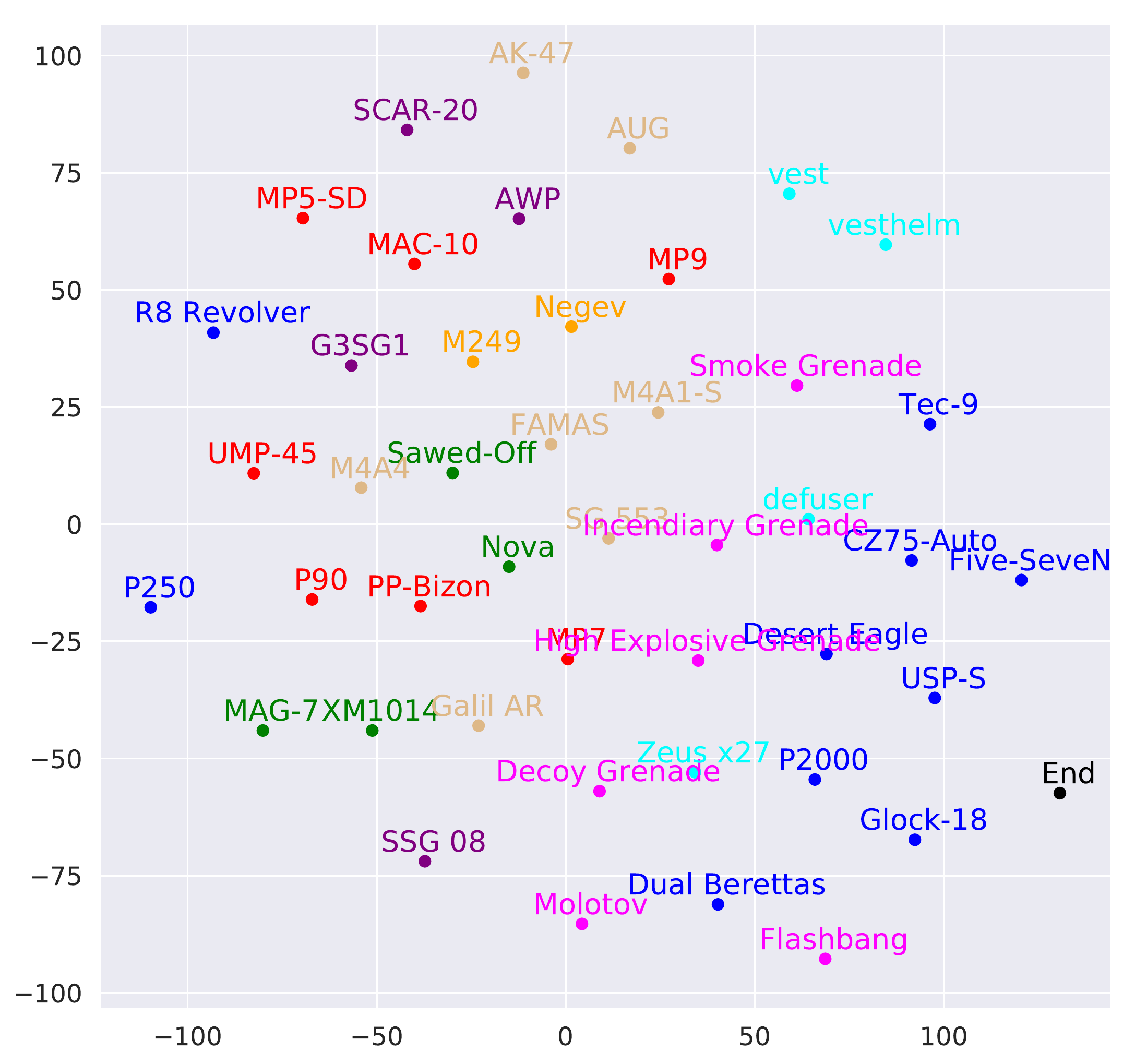}
    \caption{\textbf{Atomic action Embedding t-SNE visualization.} Different color stands for purchase different types of weapons. We further categorize guns into 6 types just for visualization.}
    %pistols, shotguns, submachine gun, rifles, machine guns 
    \label{fig:tsne}
 \end{figure}{}

%In this section, we introduce (1) our Model Agnostic Meta-Learning algorithm based on Reptile \cite{nichol2018reptile}, (2) our weapon embeddings are pre-trained using word2vec \cite{mikolov2013distributed}, (3) our state encoder, (4) our multi-task decoder for action generation with strongly supervised gate network, (5) learning objective, and (6) evaluation metrics.

\subsection{Meta-learning Algorithm}

The detailed algorithm is described in Algorithm \ref{alg}. We have investigated MAML and its first-order simplification \cite{finn2017model} and later moved to Reptile \cite{nichol2018reptile} since it is also first-order and performs reasonably well in many tasks. Note that the original algorithm is used for image classification and each task/class contains several data. In our case, however, we have more tasks/games and fewer data to support. Thus, for target sets of training data, we update the model parameters as well while the original approach performs evaluations only after several epochs. For the meta-learning loop, we use vanilla SGD that samples a single task for each step. In the inner loop for adapting to each task, we use Adam \cite{kingma2014adam} as optimizer.

\begin{algorithm}[t]
   \caption{Our modified model agnostic reptile meta-learning algorithm}
   \label{alg}
\begin{algorithmic}
  \STATE $k = $ number of shots in few-shot learning.
  \STATE $\epsilon = $ meta-learning step size.
   \STATE Initialize model parameters \textsc{$\theta$}.
   \REPEAT
   %\FOR{$i = 1, 2, ...$}
   \STATE Sample single match data \textsc{$M_i$} with repetition
   \FOR{\textsc{$j$} = 1 {\bfseries to} $k$}
   \STATE Sample single round \textsc{$R_{ij}$} and compute loss \textsc{$L$}
   \STATE \textsc{$\theta'$} $\leftarrow$ \textsc{$\theta$} $-$ $\nabla_\theta L$
   \ENDFOR
   \STATE Compute target set loss \textsc{$L'$} on the other rounds.
   \STATE \textsc{$\theta''$} $\leftarrow$ \textsc{$\theta'$} $-$ $\nabla_{\theta'} L'$
   \STATE Update \textsc{$\theta$} $\leftarrow$ \textsc{$\theta$} $+$ $\epsilon (\theta''-\theta)$
   \UNTIL{convergence}
\end{algorithmic}
\end{algorithm}

\subsection{Atomic Action and Embedding}
Although each round only requires a one-time purchase to receive the result, the purchasing action space is huge and complex. It contains different combinations of 44 weapons for each player. %with a maximum of purchasing 12 weapons at the same time.
We thus approximate the complex joint action by splitting it into a sequence of \textbf{\textit{atomic actions}} and each atomic action represents a single weapon purchase.
According authors' human priors, professional players mostly show the same pattern of purchasing weapons: they prefer to buy guns first, then grenades, and equipment at last if there are still money left. 
We therefore sort the action sequence by the specific order in accordance with the professional purchasing habit. Then we formulate the one-time purchasing action into a sequence of atomic actions and consider it as a sequence generation problem.

Therefore we can narrow down the action space. We pre-train the atomic action embeddings of purchasing each weapon by continuous bag-of-words model \cite{mikolov2013efficient} using the context information in the training set. 
% The t-SNE visualization of generated embedding is shown in Figure \ref{fig:tsne}. 
By visually inspecting the t-SNE of atomic action embeddings (Figure~\ref{fig:tsne}), we verify that purchasing similar weapons are more close to each other which can support the sequence decoding process. We highlight guns in different colors based on its subtype. The "End" action will terminate the purchase.
%: pistols, shotguns, sub-machine guns, rifles, machine guns and others. 

%The action embeddings are pretrained using the word2vec \cite{mikolov2013distributed} on training data. \Deren{Not clear. Cite t-SNE graph. describe how we convert it into a sequence to sequence problem. Then move to multitask seq2seq.}
%The embedding are generated by gun attributes. For equipment that do not have gun attributes such as vest, grenade, end token, we fill in the attribute if applicable and append an one-hot representation at the end of the embedding for each different equipment. The visualization of generated embedding is shown in Figure \ref{fig:tsne}. The attributes are later normalized between $[0,1]$ during training time. The embedding can be found in our source code.

\subsection{State Encoder} 

The state of the agent includes the weapons of itself, teammates and enemies. We use hierarchical attention \cite{yang2016hierarchical} to represent the aggregated representation of each player and each team. By using hierarchical attention, we fuse the order-independent weapons representations into player representations and player representations into team representations. With the round attribute encoder, we leverage the round information by incorporating the weapon history and performance score of the agent at past rounds to force the agent to learn from the past rounds.

\subsubsection{Weapon Encoder}
Not all weapons contribute equally to the representation of the player's attribute. Hence, we apply the attention mechanism to attend to more important weapons for the player and aggregate the representation of the weapons to vectorize a player. Specifically, weapon representation $p_i$ for player $i$ is
\begin{equation}
u_{i,t} = tanh(W_1 x_{i,t} + b_1),
\end{equation}
\begin{equation}
\alpha_{i,t} = \frac{exp(v_1^T u_{i,t})}{\sum_t exp(v_1^T u_{i,t})},
\end{equation}
\begin{equation}
p_i = \sum_t \alpha_{i, t} x_{i,t}.
\end{equation}
That is, we get the hidden representation $u_{i,t}$ from weapon embedding $x_{i,t}$ of $t$-th weapon of player $i$. We then measure the weight of the weapon $\alpha_{i,t}$ based on the similarity between $u_{i,t}$ and a weapon importance vector $v_1^T$. After that, the aggregated representation of weapons of one player is the weighted sum of the weapon embeddings. $W_1$, $b_1$ and $v_1^T$ are trainable parameters, and shared for all players.

%We then get the final player representation by concatenate $z_i$ with player money $y_i$ and player performance score $r_i$:
%\begin{equation}
%p_i = [z_i; y_i; r_i].
%\end{equation}

\subsubsection{Team Encoder}
Similarly, we use the attention mechanism to assign weights to players in each team and thereby compute the aggregated team representation $z$.
% \begin{equation}
% u_i = tanh(W_2 p_i + b_2),
% \end{equation}
% \begin{equation}
% \alpha_i = \frac{exp(v_2^T u_i)}{\sum_i exp(v_2^T u_i)},
% \end{equation}
% \begin{equation}
% z = \sum_i \alpha_i p_i.
% \end{equation}
% $W_2$, $b_2$ and $v_2^T$ are trainable parameters, and shared for two teams, i.e., allies and enemies. 
%Note that to represent state of agent itself, allies and enemies, we just need to compute team representation for allies and enemies, and agent's hidden representation do not go through this team-level attention.

\subsubsection{Round Attribute Encoder}
In the proposed round-based task, previous rounds' actions and feedbacks are explicitly perceived and should be carefully reflected by the agent. Therefore, round information is a key feature that needs to be effectively utilized. Specifically, in the $j$-th round, the agent knows its final weapons after purchasing from previous rounds \{$X'_{1}, \dots, X'_{j-1}$\} and performance scores \{$s_{1}, ..., s_{j-1}$\}. The past weapons are passed into weapon encoder to get the aggregated weapon representation of previous rounds. Then we normalize the performance scores as weights and compute the weighted sum of past weapon representation as round representation $h_r$. As a result, the agent will attend to rounds in which it has better performance.

\subsubsection{Economy Encoder}
To take the economy into account, we encode the normalized money features of all players through a multi-layer perceptron (MLP) as a dense economy vector $h_c$.

\subsubsection{State Representation}
We concatenate the agent's player representation $p^s$, two teams' representation $z^a$, $z^e$, round representation $h_r$ and money representation $h_c$. Then we get the overall initial state representation using MLP.
%Then we get the overall state encoding by concatenating $z$ with team side $d$ and match score $g$:
\begin{equation}
h = W_{h2}\ ReLU(W_{h1} [p^s; z^a; z^e; h_r; h_c]).
\end{equation}
%where $p^s$ is the agent's player representation, and $z^a$ and $z^e$ is the team representation for allies and enemies, respectively.

\subsection{Multi-Task Decoder}
Given the player's states, the agent is asked to take sequence of atomic actions and receive a reward from the environment. Such atomic actions can be classified into three categories by their types, i.e. purchasing guns, grenades or equipment. Since attribute of each type of weapon and the prices of weapons are diverse, the strategies for generating different types of weapons should be different. We formulate the purchasing as a multi-task atomic action sequence generation problem.

\subsubsection{Gate Network}
Before the decoding step, we train three gate networks to control the atomic action generation of each task. Each gate is a binary classifier, a simple MLP, which decides whether to generate actions of a task. They are trained independently throughout the entire training procedure with strong supervision signals: whether a label has actions of this task. The gates can facilitate action generation and are easy to train.

%The gates can facilitate the action generation and are easy to train because it receives strong supervision signal.

\subsubsection{Task-Specific Decoder}
Based on the initial state representation, the agent generates atomic actions sequentially and transits state using LSTM \cite{hochreiter1997long}. Each LSTM-based decoder is designed for one task and all decoders share the same player money information and the initial state representation given by the encoder. This multi-task design also ensures a better generalization of the encoder with training signals of different domains \cite{caruana1997multitask}.

The agent takes the state representation $h$ to initialize the LSTM hidden state $h_0$ and use the hidden state $h_t$ at each time step $t$ to generate the distribution over atomic actions.
\begin{equation}
h_t = LSTM(h_{t-1}, a_{t-1}),
\end{equation}
\begin{equation}
%P_{t} = softmax(W_{c2}\ ReLU(W_{c1}h_t)).
%d_{t} = softmax(W_{c2}\ ReLU(W_{c1}h_t)).
P(\cdot|a_1, ..., a_{t-1}, h) = \sigma(W_{c2}\ ReLU(W_{c1}h_t)).
\end{equation}
where $\sigma$ is the softmax function.

\subsection{Learning Objective}
\label{subsec:objective}
Since we use LSTM to generate the atomic actions sequentially, using the "teacher forcing" algorithm \cite{williams1989learning} to train our model will inevitably result in the exposure bias problem \cite{ranzato2015sequence}:  maximizing the likelihood of a sequence of atomic actions needs the ground truth atomic action sequence during training but such supervision signal is not available in testing, thus errors are accumulated while generating the atomic action sequence. 
%Besides, since the order of atomic actions does not matter in this task, forcing the agent to generate atomic actions at an assigned order is harmful. 
To address the issue, we use the self-critical sequence training (SCST) method \cite{rennie2017self}.
SCST is a form of REINFORCE ~\cite{williams1992simple} algorithm that is designed for tackling sequence generation as RL problem. 
We first sample an atomic action $a^s_t \sim P(\cdot|a^s_1, ..., a^s_{t-1}, h)$ from the atomic action distribution at each generating step $t$ to get an atomic action sequence $A^s$. 
%This is the stochastic inference process of our model to solve the Markov Decision Process (MDP) in the reinforcement learning literature. 
Another sequence $A^g$ is generated using greedy search to maximize the output probability distribution $P(\cdot|a^g_1, ..., a^g_{t-1}, h)$ at each step and serves as a baseline. We define $r(A)$ as the reward function. We compute $F1$ score with ground-truth as our reward function since we do not consider the atomic action sequence order and compare the formed atomic action set only. The objective function is defined as follows.

\begin{equation}
L = (r(A^g)-r(A^s))\sum_t log\ p(a_t^s|a_1^s, ..., a_{t-1}^s).
\end{equation}

Our gate network is trained separately with strong supervision signal. We compute the cross entropy loss for each binary gate.

\begin{table*}[ht!]
\centering
\scalebox{0.90}{
\begin{tabular}{l | cccc}
\toprule
\textbf{Method}          & \textbf{$\boldsymbol{F_1}$} & \textbf{$\boldsymbol{F_1}$-gun} & \textbf{$\boldsymbol{F_1}$-grenade} & \textbf{$\boldsymbol{F_1}$-equip} \\
\midrule
Greedy Algorithm     & 0.2612 & 0.0338 & 0.3831 & 0.2890\\
\midrule
Single-Sequence Reasoner w/ Gate     & 0.5109   & 0.3487       & 0.5734      & 0.7119  \\
Single-Sequence Reasoner + RAE w/ Gate    & 0.5206   & 0.3494       & 0.5870      & \textbf{0.7177}  \\
\midrule
Multi-Sequence Reasoner w/o Gate   & 0.5028   & 0.3216       & 0.5912      & 0.5370  \\
Multi-Sequence Reasoner + RAE w/o Gate   & 0.5114   & 0.3216       & 0.6008      & 0.5435 \\
\midrule
Multi-Sequence Reasoner w/ Gate            & 0.5475   & \textbf{0.4524}       & 0.6608      & 0.5386  \\
Multi-Sequence Reasoner + RAE w/ Gate & \textbf{0.5670}   & 0.3920       & \textbf{0.6639}      & 0.6006  \\
\bottomrule
\end{tabular}
}
\caption{\textbf{The results of different methods including ablation study.} We categorize weapon outputs by gun, grenade and equipment (equip) to get more insights. The multi-sequence reasoner with round attribute encoder (RAE) and gate classifier achieves the best result. Both gate classifier and RAE can improve model performance in different circumstances.}
\label{tab:result}
\end{table*}

\subsection{Evaluation Metrics}
We evaluate the methods by calculating the $\mathbf{F_1}$ \textbf{score} between the model output atomic action sequence and ground truth. Same as the reward function used in learning objective. 
%Given model output atomic action sequence $A$ and ground truth atomic action sequence $\hat{A}$, precision and recall are defined as:
% \begin{equation}
%     Precision = \frac{|A \cap \hat{A}|}{|A|},
% \end{equation}
% \begin{equation}
%      Recall = \frac{|A \cap \hat{A}|}{|\hat{A}|},
% \end{equation}

% \begin{equation}
%     F_1 = 2 \cdot \frac{Precision \cdot Recall}{Precision + Recall}.
% \end{equation}{}

\section{Experiments}

In this section, we evaluate the proposed methods on the test set based on their highest performance on the development set. During atomic action generation step, we mask out weapons that cannot currently afford all methods to avoid invalid purchases in both training and testing phase. In addition, some weapons such as grenades has a quantity limit. We mask out weapons that currently reach the quantity limit as well.

We do an ablation study on our multi-sequence reasoner to measure its effectiveness.% The results are reported in Table~\ref{tab:result}. 

\subsection{Greedy Algorithm Baseline}
The greedy method buys weapons based on type order. For each type, it will do a sequence of purchases prioritized by weapon expensiveness. In other words, at each purchasing step, it buys the most expensive weapon which is affordable in that type. It only buys one gun, a maximum number of grenades up to the quantity limit, and then buy equipment. We consider it as the baseline as it does not require training and it is hard to generalize to different gaming scenarios. %From Table~\ref{tab:result}, it has the lowest $F_{1}$ score in total. However, it has a strong $F_{1}$ score for guns. This is partly because most people only buy one gun which is not ideal for sequence generation. 

\subsection{Multi-Sequence Reasoner}
The Multi-Sequence Reasoner follows the model architecture described in section \ref{method}. In the first round, all players start with no weapons and are restricted to buy pistols due to finance issues. Data in the first round are not generalizable and we cannot utilize its information for the second round. Since we need useful empirical information of past rounds and the second round only contains data from the first round, these two rounds are removed. In the 16th round, the two teams switch sides and start from scratch. Therefore 16th and 17th rounds are not included in our task as well.

We set the few shot number $K$ to 5. We set the batch size to 10 to generate the atomic actions of ten players independently in a match at the same time. Note that we tackle this task as a single agent problem and leave team-based multi-agent purchasing prediction to future work. During inference time, we use the beam search to generate optimal atomic actions and set the beam size to 1. 

\subsubsection{Round Attribute Encoder}
We evaluate the effectiveness of round attribute encoder by concatenating its output to the state representation encoded by the original sequence reasoner without modifying the original model architecture. To measure its effectiveness, for all experiments, we run the model with two settings: with and without the round attribute encoder.

\subsection{Result}
We report the performance of different methods in Table~\ref{tab:result}. We also showed the performance of each type of weapon purchasing. First of all, we observe that the naive Greedy Algorithm does not achieve a good performance comparing to deep learning models. Besides, we also observe that multi-sequence reasoner with round attribute encoder in the last row of table~\ref{tab:result} achieves the highest $F_{1}$.

\subsubsection{Ablation Study}
To test the importance of the gate network (Gate), round attribute encoder (RAE), and multi-task decoder (Multi-Sequence Reasoner), we perform an ablation study where we remove the gate network, round attribute encoder, and turn multi-task decoder into single decoder (Single-Sequence Reasoner). As shown in Table~\ref{tab:result}, ablating gate network, round attribute encoder, and multi-task decoder from our integrated model will impair the performance and lead to a decrease of 5.56\%, 1.95\% and 4.64\% for $F_1$ score. More importantly, consistently decreased performance in all three ablation models due to RAE removal shows the importance of utilizing round meta information. Thus, we believe round-based games are fundamentally different from conventionally studied continuous games. How to learn an effective round meta information representation and how to utilize it is an important topic for future game AI studies.

\section{Conclusion}
We explored the challenges in round-based games in which each data contains a long sequence of dependent episodes. We introduced a new round-based dataset based on CS:GO. The dynamic environment and connections between rounds make it suitable for round-based game study. We presented a few-shot learning task to encourage the agent to learn general policies and can quickly adapt to players' personal preferences in certain scenarios. Experimentally, we showed that our proposed model, Multi-Sequence Reasoner, is effective. We found that using round empirical information leads to nontrivial improvement on the result, thus testifying the importance of round history for the task. We believe our research will open doors for building interpretable AI for understanding episodic and long-term behavioral strategies not only for the gaming community but also for the broader online platforms.

%In each game with ten new players, the agent is required to learn to purchase weapons given only a few shots of round experience. Experimentally, we showed that our Multi-Sequence Reasoner is effective. We found that simply using round empirical information leads to nontrivial improvement on the result, thus testifying the importance of round history for the task. We believe our research will open doors for building interpretable AI for understanding episodic purchasing decision not only for gaming community but also for the broader online platforms.

%As future work, we plan to do the following: (1) Explore the more challenging multi-agent task for our dataset to train collaboration purchase as a player can purchase certain weapons for teammates. (2) Study the use of round information in depth. (3) Look into how round-based information can be applied to continuous environments by dividing them into micro-games. 

\section{Acknowledgements}
The authors would like to thank Yang Yang, Sijie Duan and Qi Luo for their help on dataset collection. YZ and EF are grateful to DARPA (grant no. D16AP00115).

% \textbf{Acknowledgements}. YZ and EF are grateful to DARPA (grant no. D16AP00115).

\bibliography{main}
\bibliographystyle{aaai}
\end{document}